\documentclass[11pt,a4paper]{article}
\usepackage[margin=1in]{geometry}
\usepackage{setspace}
\usepackage[utf8]{inputenc}
\usepackage[T1]{fontenc}
\usepackage{lmodern}
\usepackage{amsmath,amssymb,amsthm,mathtools}
\usepackage{graphicx}
\usepackage{booktabs}
\usepackage{tabularx}
\usepackage{enumitem}
\usepackage{placeins}
\usepackage{microtype}
\usepackage{natbib}
\bibpunct{(}{)}{;}{a}{,}{,}
\usepackage{etoolbox}
\usepackage{xcolor}
\AtBeginEnvironment{table}{}
\usepackage[colorlinks=true,linkcolor={blue!60!black},citecolor={green!45!black},urlcolor={blue!70!black}]{hyperref}
\usepackage{fancyhdr}

\DeclareMathOperator*{\argmax}{arg\,max}
\providecommand{\doi}[1]{\href{https://doi.org/#1}{\nolinkurl{doi:#1}}}
\newcommand{\E}{\mathbb{E}}
\newcommand{\filt}{\mathcal{F}}
\newcommand{\Nb}{\mathcal{N}}
\newcommand{\Pk}{P_k}
\newtheorem{theorem}{Theorem}
\newtheorem{proposition}{Proposition}
\newtheorem{corollary}{Corollary}
\theoremstyle{definition}
\newtheorem{assumption}{Assumption}

\title{\textbf{\LARGE When Does Design Thinking Work?\\[0.3em] \Large A Convergence Theorem for the Iterate--Test Core}\\[0.6em]
\normalsize\textit{Under review at Journal of Product Innovation Management (JPIM)}}

\author{
  \textbf{Pawe{\l} Kura{\'s}}$^{1,*}$, \quad
  \textbf{Adrian Michalski}$^{2}$, \quad
  \textbf{Alicja Gerka}$^{3}$, \quad
  \textbf{Patryk Organi{\'s}ciak}$^{3}$\\[0.8em]
  $^{1}$\small Institute of Informatics, Kozminski University, ul.~Jagiello{\'n}ska 57/59, 03-301 Warsaw, Poland\\
  $^{2}$\small Faculty of Mathematics and Applied Sciences, Rzesz{\'o}w University of Technology,\\ \small al.~Powsta{\'n}c{\'o}w Warszawy 12, 35-959 Rzesz{\'o}w, Poland\\
  $^{3}$\small Department of Complex Systems, Rzesz{\'o}w University of Technology,\\ \small al.~Powsta{\'n}c{\'o}w Warszawy 12, 35-959 Rzesz{\'o}w, Poland\\[0.5em]
  $^{*}$Corresponding author: \texttt{pkuras@kozminski.edu.pl}
}
\date{\small\today}

\begin{document}
\maketitle
\thispagestyle{fancy}

\begin{abstract}
\noindent\textbf{Academic Summary:} Design thinking is widely taught as an iterative loop of ideation, prototyping, and testing, yet conditions under which such a loop must improve innovation outcomes remain unspecified. Taking problem framing as given, we formalize the iterate--test core as search over a finite prototype space whose utility is learned through noisy evaluations. We prove that the prototype a team believes to be best converges almost surely to a local optimum under five conditions: finite framing, local ideation, non-summable discovery rates, unbiased repeated testing, and retention of tested prototypes. Bounded test bias or criteria drift degrades the guarantee by a bounded amount, and multiple stakeholders yield convergence to a locally Pareto-optimal set. Finite-horizon simulations across three landscape classes (additive valuation, NK rugged landscapes, and clustered modular architectures) show that asymptotic limits are rarely reached in practice. Teams face an optimizer’s curse, overestimating their chosen concept by 4--8\% and frequently discarding superior prototypes. Testing every prototype once proves to be the worst allocation; optimal performance requires building one to three prototypes per five tests and repeatedly re-testing leaders. Pre-screening concepts yields greater performance gains than tuning testing policies.

\vspace{0.8em}
\noindent\textbf{Managerial Summary:} While design thinking is celebrated for encouraging rapid prototyping and user feedback, product managers lack clear guidelines on how to budget and manage iterative testing. This study models the prototyping cycle to discover how innovation teams can achieve the best possible product under realistic time and budget constraints. Our findings provide four practical rules for innovation leaders: First, never test every prototype just once. Showing many prototypes to users once is the worst use of a development budget; teams should instead build one to three new concepts for every five tests and allocate the remaining tests to verify leading concepts. Second, discount initial enthusiasm: teams systematically overestimate the value of their favorite concept by 4\% to 8\% due to selection bias. Third, invest in rough pre-screening of ideas before building physical prototypes, which delivers greater performance gains than sophisticated testing algorithms. Fourth, confirmation bias toward a favorite design causes little harm as long as it remains smaller than test measurement noise. Finally, managers should schedule at least one deliberate return to the problem definition, as iterative testing alone will not signal when a fundamental reframe is needed.

\vspace{0.8em}
\end{abstract}

\vspace{1.0em}

\section{Introduction}

Design thinking asks teams to understand the people they are designing for, frame a problem, generate many ideas, build cheap prototypes and test them with users, and to repeat the last three steps until something works \citep{dschool2010}. The method has spread from product design into management, public policy and health care \citep{liedtka2015, mintrom2016, altman2018} without a theory of why it should work. Reviews describe a construct that is fragmented and thinly grounded \citep{johansson2013, micheli2019}, and in health care, where the evidence has been reviewed systematically, it consists mostly of small studies of fair or poor quality \citep{altman2018}.

This paper asks a narrow question that can be answered exactly. \emph{Under what conditions does the iterate--test core of design thinking provably improve what a team builds, and what does the guarantee leave out?} We take the framing of the problem as given and study the loop that follows it: propose variants, build one, test it, revise what is believed about every prototype built so far, and repeat. A prototype is a set of features. Its value is a utility that the team can observe only through noisy tests, and the team keeps a belief about each prototype it has tested. A local optimum is a prototype that no nearby variant beats.

The first part of the paper is a convergence result. Suppose the team entertains finitely many features, generates ideas near what it has built, keeps producing improving ideas at a rate that is not summable, tests without bias and re-tests what it has kept, and discards nothing it has tested. Then, with probability one, the prototype the team believes to be best is eventually a fixed one, and it is a local optimum (Theorem~\ref{thm:converge}). Results of this kind are known for search that retains its best solution and for stochastic search under noise. We state the result for the design loop, where each condition names something a team does and can be weakened to see what is lost: a bias of at most $B$ in the tests leaves the team at a prototype that no variant beats by more than $2B$, criteria that drift within a band of half-width $D$ behave like a bias of size $D$, and when stakeholders disagree the set of non-dominated prototypes converges to a locally Pareto-optimal set.

The second part measures what the theorem leaves open, by simulation on three classes of landscape whose local optima are known exactly. The limit turns out to be remote, so what a team has when it stops depends on how fast the loop climbs. A team's choice among its own prototypes costs it about as much as its failure to find better ones, and its estimate of the prototype it has chosen is too high. Under a fixed budget of tests, never testing anything twice is the worst policy. A bias in favour of the current favourite does no measurable harm until it is about as large as the noise of a test. We also compare the loop with bandit and local-search policies, and teams that reframe with teams that do not. Section~\ref{sec:empirical} sets out what would have to be measured to learn whether design teams satisfy the conditions.

\section{Background and related work}
\label{sec:related}

The five modes of the Stanford d.school (empathise, define, ideate, prototype, test) \citep{dschool2010} and the Design Council's Double Diamond \citep{designcouncil2007} are two widely taught descriptions, and Brown adds the criteria by which an outcome is judged: what people want, what can be built, and what a business can sustain \citep{brown2008, brown2009}. Experimentation and iteration recur in accounts of the method \citep{carlgren2016, micheli2019}. \citet{liedtka2015} proposes that it improves outcomes by mitigating biases of judgement, among them the tendency to seek confirmation for a solution one already favours; designers shown an example solution are known to reproduce its features, flaws included \citep{jansson1991}. Novice teams differ in how they combine need-finding, brainstorming and prototyping \citep{seidel2013}; designers who built several prototypes before receiving feedback produced better designs than those who refined one at a time \citep{dow2010}; and groups that generate more ideas do not thereby select better ones \citep{rietzschel2006}.

Formal theories of design begin with Simon, for whom design is search under bounded rationality that ends in a satisfactory alternative \citep{simon1956, simon1996}. Later accounts describe what design reasoning operates on: function, behaviour and structure \citep{gero2004}, the joint expansion of concepts and knowledge \citep{hatchuel2009}, the mapping from requirements to parameters \citep{suh1990}, and the coupled tasks that force rework \citep{smitheppinger1997}. Problem and solution are held to evolve together \citep{maherpoon1996, dorstcross2001}. Problem setting is central to professional practice \citep{schon1983}, framing is an abductive step \citep{dorst2011}, and experienced designers routinely reframe the briefs they are given \citep{paton2011}. \citet{mccomb2015} simulate a design team as agents performing simulated annealing and compare its behaviour with human data. None of these accounts states conditions under which an iterative process must improve its outcome. Our model is the simplest representation in which that question can be posed: design as exploration of a space of alternatives \citep{woodbury2006}, with the frame held fixed.

Outside design, convergence of search to a local optimum is well understood. Adaptive walks on rugged landscapes end at local optima \citep{kauffmanlevin1987}, and in the NK model the optima multiply and the walks shorten as interactions among components increase \citep{kauffmanweinberger1989, kauffman1993}. Organisational search of this kind adapts well locally and misses distant peaks \citep{levinthal1997, levinthalmarch1993, march1991}. A canonical genetic algorithm does not converge to the global optimum unless it always retains the best solution found \citep{rudolph1994}. In design a prototype's value is estimated, and search under noisy evaluation has its own literature: simulated annealing converges under noise that is driven down fast enough \citep{gutjahr1996}, random search over a discrete set converges when samples grow at a suitable rate \citep{homem2003}, evolutionary algorithms re-sample candidates \citep{jin2005}, and in ranking and selection an efficient budget gives most replications to the leading alternative and its close competitors \citep{chen2000}. Imperfect evaluation changes organisational search and can even help it leave a local peak \citep{knudsen2007}. The economics of testing in product development covers the choice of test mode \citep{thomke1998}, parallel and sequential testing \citep{loch2001}, the number of concepts to test \citep{dahan2001} and correlated designs \citep{erat2008}.

A team that maintains beliefs about alternatives and chooses what to test next is also solving a problem studied in Bayesian optimisation \citep{jones1998, shahriari2016}, with expected improvement and the knowledge gradient \citep{frazier2008, frazier2009} as standard criteria and regret bounds for smooth objectives \citep{srinivas2012}. In a finite space with independent beliefs the closer analogy is best-arm identification, where the number of tests needed to separate two alternatives grows, up to logarithmic factors, with the inverse square of the difference between them \citep{evendar2006, mannor2004, audibert2010}. These results explain why choosing among prototypes is slow and supply the policies with which we compare the loop \citep{auer2002, russo2018}.

\section{The model}
\label{sec:model}

\begin{figure}[t]
\centering
\includegraphics[width=0.72\textwidth]{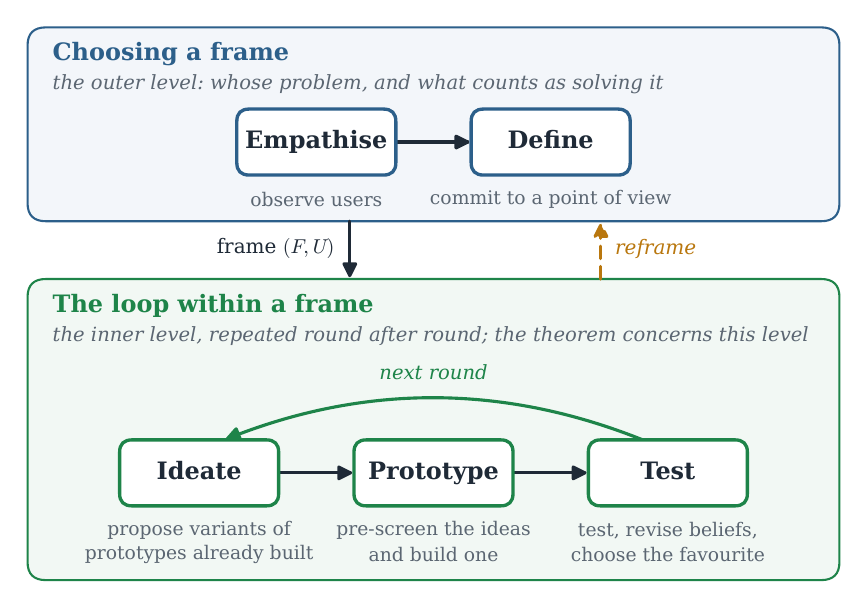}
\caption{The two levels of the model. Empathising and defining produce a frame, a set of features $F$ and a utility $U$. Within the frame the team repeats a round of ideation, prototyping and testing, and the theorem concerns this loop. A reframe (dashed) returns the team to the outer level and starts a new framing epoch.}
\label{fig:process}
\end{figure}

The model has two levels, the choice of a frame and the loop that runs inside it (Figure~\ref{fig:process}).

\subsection{Frames, prototypes and the loop}

The outer level is the choice of a \emph{frame}. Empathising gathers observations about users, and defining commits the team to a point of view, a statement of whose problem is being solved and what would count as solving it. In Dorst's account this step is abductive: starting from the value it wants to create, the team conjectures both the thing to be made and the working principle by which that thing would deliver the value \citep{dorst2011, kolko2010}. We do not model how the conjecture is produced. Its result, for our purposes, is a pair $(F,U)$: a finite set $F$ of features that solutions may combine, and a utility $U$ that says how good each combination is. A \emph{reframe} replaces $(F,U)$ by another pair, and a \emph{framing epoch} is the stretch of the project between two reframes.

The inner level is the loop inside an epoch. A \emph{prototype} is a subset $p\subseteq F$, so the space of prototypes is $2^F$. In round $k$ the team proposes \emph{ideas}, each a variant of a prototype it has already built; it chooses one idea and builds it; and it tests the new prototype and some of the older ones, revises its beliefs, and so decides which prototype is currently its \emph{favourite}. 
Consider a team designing a travel mug for a commuter, with a frame that admits four features: a folding handle, a one-touch lid, a spill seal and a temperature display. If ideation changes one feature at a time, the mug with a lid and a seal is a local optimum when adding the handle or the display, or removing the lid or the seal, makes it worse, even if a mug three changes away would be better still. Deciding that the commuter is the wrong user is a reframe.

We write $\Pk\subseteq 2^F$ for the set of prototypes built and tested by the end of round $k$, and $\filt_k$ for the history up to that point: the ideas proposed, any rough estimates made of them, the prototypes built, and the tests and their results (Table~\ref{tab:notation}).

\begin{table}[t]
\caption{Notation.}
\label{tab:notation}
\centering\small
\begin{tabularx}{\textwidth}{@{}lX@{}}
\toprule
$F$, $2^F$ & features admitted by the frame; prototype space\\
$U(p)$, $U^*$ & utility of prototype $p$, in $[0,1]$; its maximum\\
$\Nb(p)$ & variants of $p$ that ideation can reach in one step\\
$\Pk$, $P_\infty$ & prototypes built and tested by round $k$; their eventual set\\
$\filt_k$ & history up to the end of round $k$\\
$M^p_k$ & belief about $p$ after round $k$ ($\widehat M^p_k$ under biased tests)\\
$p^*_k$, $X_k$ & the favourite, $\argmax_{p\in \Pk}M^p_k$, and the belief about it\\
$\delta_k$ & lower bound on the chance that round $k$ builds a better variant of the favourite\\
$\sigma$ & standard deviation of the noise of a test\\
$b_k$, $B$, $b$ & bias of the test in round $k$; its bound; the constant bias used in simulations\\
\bottomrule
\end{tabularx}
\end{table}

\subsection{Utility, tests and beliefs}

Following \citet{brown2009}, a solution is judged by its desirability to users, its technical feasibility and its business viability. In our examples the utility is a weighted sum of the three,
\begin{equation}
\label{eq:utility}
U(p)=w_{\mathrm d}U_{\mathrm d}(p)+w_{\mathrm f}U_{\mathrm f}(p)+w_{\mathrm v}U_{\mathrm v}(p),
\end{equation}
with components in $[0,1]$ and weights summing to one \citep{keeney1976}. The theory uses only two properties of $U$: it is bounded, and it does not change during an epoch.

The team does not know $U$. A test of prototype $p$ in round $k$ returns
\begin{equation}
\label{eq:test}
y=U(p)+b_k+\varepsilon_k ,
\end{equation}
where $\varepsilon_k$ has mean zero and variance at most $\sigma^2$ given the history and the choice of what to test, and $b_k$ is a systematic error, possibly depending on the prototype and the history, with $|b_k|\le B$. User interviews, expert critiques and feasibility estimates are all tests in this sense.

From its test results the team forms a belief $M^p_k$ about each prototype it has built: an estimate of its utility, such as the average of the results or the posterior mean of a Bayesian model \citep{degroot2004}. The favourite $p^*_k$ is the prototype with the highest belief, ties being broken by any fixed rule, and
\begin{equation}
\label{eq:Xk}
X_k=\max_{p\in \Pk}M^p_k
\end{equation}
is how good the team thinks its favourite is. Two other quantities must be kept apart from $X_k$. The true utility $U(p^*_k)$ of the favourite is what the team would deliver if it stopped now, the \emph{shipped} utility. The best true utility among all tested prototypes is what the team has \emph{found}, whether or not it knows it.

An idea that has not been built has no test results, so the choice of which idea to build must rest on something cheaper, such as a rough estimate, an analogy or a critique. We call this choice \emph{pre-screening}. It affects the speed of the loop but not its limit.

\section{Convergence within a frame}
\label{sec:theory}

\subsection{Five assumptions}

\begin{assumption}[Finite frame]
\label{as:finite}
Within an epoch the feature set $F$ is finite, the utility $U:2^F\to[0,1]$ does not change, and no two prototypes have exactly the same utility.
\end{assumption}

Ties are absent for almost every utility measured on a continuous scale. Finiteness is a real restriction, because it rules out the arrival of features nobody had thought of; if additions stop, the results apply from the last one onward.

\begin{assumption}[Local ideation]
\label{as:local}
Each prototype $p$ has a set $\Nb(p)\subseteq 2^F$ of variants, and every idea, apart from the ideas a team starts with, is a variant of a prototype already in $\Pk$. A prototype is \emph{locally optimal} if none of its variants has higher utility.
\end{assumption}

In the simulations variants differ from $p$ in at most two features. If every prototype were a variant of every other, the limit would be the global optimum, at a price in speed that Section~\ref{sec:ablation} measures.

\begin{assumption}[Persistent generativity]
\label{as:gen}
There are numbers $\delta_k\ge0$, determined by the history $\filt_{k-1}$, such that whenever the favourite $p^*_{k-1}$ has a better variant that has not yet been built, round $k$ builds and tests such a variant with conditional probability at least $\delta_k$ given $\filt_{k-1}$, and
\begin{equation}
\label{eq:nonsummable}
\sum_{k\ge1}\delta_k=\infty .
\end{equation}
\end{assumption}

Ideas may become harder to find near a peak: a chance of improvement that falls like $1/k$ satisfies \eqref{eq:nonsummable}, one that falls like $1/k^2$ does not. The team need not recognise the better variant when it builds it, nor know that one exists.

\begin{assumption}[Unbiased, repeated testing]
\label{as:test}
Tests are unbiased ($B=0$), every prototype in $\Pk$ is tested infinitely often, and the belief about a prototype differs from the average of its test results by an amount that vanishes as its tests accumulate.
\end{assumption}

The last clause holds for the average itself and for conjugate posterior means under any prior. Beliefs then become correct: the noise accumulated over the tests of a prototype is a martingale with increments of bounded variance, so by the strong law for martingales \citep[Sec.~4.5]{durrett2019} the average of its results converges to $U(p)$, and $M^p_k\to U(p)$ almost surely for every retained prototype, in whatever order the tests are chosen.

\begin{assumption}[Retention]
\label{as:retention}
No tested prototype is discarded: $\Pk\subseteq P_{k+1}$.
\end{assumption}

Retention keeps the best prototype found available for comparison, the role that elitism plays in evolutionary search \citep{rudolph1994}.

\subsection{The theorem}

\begin{theorem}
\label{thm:converge}
Let Assumptions~\ref{as:finite}--\ref{as:retention} hold. Then almost surely the favourite is eventually a single prototype $p_\infty$, this prototype is locally optimal, and $X_k\to U(p_\infty)$.

If, in addition, $U$ is drawn from a prior distribution on utilities with values in $[0,1]$ and the beliefs $M^p_k=\E[U(p)\mid\filt_k]$ are the posterior means under that prior, then $X_k$ is a bounded submartingale with respect to the prior: the team never expects its belief about its favourite to fall.
\end{theorem}

\begin{proof}
By retention, $\Pk$ is a non-decreasing sequence of subsets of the finite set $2^F$, so after some random round $k_0$ it equals a fixed set $P_\infty$. Every member of $P_\infty$ is tested infinitely often, so $M^p_k\to U(p)$ for each. Their utilities are distinct, so from some round on the favourite is the member $p_\infty$ with the highest utility, and $X_k\to U(p_\infty)$.

Suppose that, with positive probability, $p_\infty$ has a variant $q$ with $U(q)>U(p_\infty)$. Such a $q$ cannot belong to $P_\infty$. On this event the favourite has a better unbuilt variant in every late round, so by Assumption~\ref{as:gen} the conditional probabilities of building one sum to infinity. By L\'evy's extension of the Borel--Cantelli lemma \citep[Sec.~4.3]{durrett2019} one is then built after round $k_0$, contradicting $\Pk=P_\infty$ for $k\ge k_0$.

For the second statement, each $M^p_k$ is a martingale with values in $[0,1]$ and the set over which the maximum in \eqref{eq:Xk} is taken does not shrink, so by the conditional Jensen inequality
\[
\E[X_{k+1}\mid\filt_k]\ge\E\Big[\max_{p\in \Pk}M^p_{k+1}\,\Big|\,\filt_k\Big]\ge\max_{p\in \Pk}\E[M^p_{k+1}\mid\filt_k]=X_k .\qedhere
\]
\end{proof}

Retention and finiteness make the set of prototypes settle, repeated unbiased testing makes beliefs correct, generativity prevents the process from stopping short of a local optimum, and locality determines which optimum is meant. The first statement holds for every utility without ties and whatever the team's prior, because it uses only the fact that beliefs become correct. The second is the familiar fact that information has non-negative expected value to a Bayesian \citep{degroot2004}, and it holds only for a team whose prior is right and whose beliefs use everything in the history. A team whose prior is much more diffuse than the real spread of utilities overestimates its favourite, because the favourite is chosen for having the highest estimate and a diffuse prior does nothing to discount the luck in that estimate. This is the optimiser's curse \citep{smithwinkler2006}, and the simulated team of Section~\ref{sec:sim} is of this kind.

\subsection{Biased tests and drifting criteria}
\label{sec:bias}

People seek and read evidence in ways that favour what they already believe \citep{nickerson1998}, and a team with a favourite is no exception. Suppose the bias in \eqref{eq:test} is bounded by $B$ and the team, unaware of it, forms its beliefs $\widehat M^p_k$ as if tests were unbiased.

\begin{proposition}
\label{prop:bias}
Let Assumptions~\ref{as:finite}, \ref{as:local} and~\ref{as:retention} hold, let every retained prototype be tested infinitely often with $|b_k|\le B$, let beliefs be averages of test results or conjugate normal posterior means, and let Assumption~\ref{as:gen} hold with $\delta_k\ge\delta$ for a constant $\delta>0$. Then almost surely, from some round on, every favourite is $2B$-locally optimal: none of its variants has a utility exceeding its own by more than $2B$.
\end{proposition}

\begin{proof}
A belief is a weighted average of the prior mean and the test results, with the weight of the prior tending to zero. Shifting each result by at most $B$ shifts the average by at most $B$, and the unbiased part converges to $U(p)$ as before, so $\limsup_k|\widehat M^p_k-U(p)|\le B$ for every $p\in P_\infty$. Let $u$ be the highest utility in $P_\infty$. A member with utility below $u-2B$ has beliefs eventually below those of the best member, so every late favourite has utility at least $u-2B$. The favourite may keep changing among such members, which is why a constant $\delta$ is assumed. Each time the favourite has a better variant not yet built, one is built with probability at least $\delta$; if this happened infinitely often, infinitely many prototypes would be built, which is impossible in a finite space. Eventually, then, no favourite has a better unbuilt variant, and every variant that has been built has utility at most $u$, which exceeds the favourite's by at most $2B$.
\end{proof}

The bound is a worst case in which the bias always works against the better of two close prototypes. A bias that merely flatters the favourite is harmful only when it is large enough to overturn a comparison (Section~\ref{sec:ablation}).

The same argument covers criteria that move. Suppose that during an epoch the utility is $U_k$ at round $k$, for instance because the weights in \eqref{eq:utility} are renegotiated, and that it stays within a band around a reference utility $\bar U$ without ties.

\begin{corollary}
\label{cor:drift}
If $|U_k(p)-\bar U(p)|\le D$ for all $k$ and $p$, then under the conditions of Proposition~\ref{prop:bias}, with unbiased tests of the current utility and with ``better variant'' understood with respect to $\bar U$, every late favourite is $2D$-locally optimal for $\bar U$ and $4D$-locally optimal for every $U_k$.
\end{corollary}

\begin{proof}
A test of $U_k(p)$ is a test of $\bar U(p)$ with a bias of at most $D$, so Proposition~\ref{prop:bias} applies with $B=D$. For a late favourite $p$ and a variant $q$, $U_k(q)-U_k(p)\le\bar U(q)-\bar U(p)+2D\le4D$.
\end{proof}

The model thus tolerates a problem that shifts within bounds while the solution is developed, which is as far as it goes towards the co-evolution of problem and solution \citep{dorstcross2001}. A discontinuous change of criteria is a reframe.

\subsection{Several stakeholders}
\label{sec:pareto}

When the weights in \eqref{eq:utility} have not been agreed, let each stakeholder $s$ in a finite set have a utility $U^{(s)}$, say that $q$ \emph{dominates} $p$ if it is at least as good for every stakeholder and better for one, and let each test return one noisy observation per stakeholder. In place of a favourite the team keeps an \emph{archive}: the tested prototypes that no other tested prototype dominates according to current beliefs.

\begin{proposition}
\label{prop:pareto}
Let Assumptions~\ref{as:finite}, \ref{as:local}, \ref{as:test} and~\ref{as:retention} hold for every stakeholder, with no ties in any $U^{(s)}$. Suppose that whenever a member of the archive has a dominating variant that has not been built, round $k$ builds such a variant with conditional probability at least $\delta_k$, with $\delta_k$ as in Assumption~\ref{as:gen}. Then almost surely the archive is eventually constant, it is the set of non-dominated members of $P_\infty$, and none of its members has a dominating variant.
\end{proposition}

\begin{proof}
As before $\Pk=P_\infty$ from some round on and all beliefs converge to the true utilities. With no ties, dominance according to beliefs eventually agrees with true dominance on $P_\infty$, so the archive settles on its non-dominated members. If a member $p$ had a dominating variant $q$, then $q\notin P_\infty$, the generativity condition would apply in every late round, and a dominating variant would be built, contradicting $\Pk=P_\infty$.
\end{proof}

With several stakeholders the loop does not resolve the disagreement: it delivers a set of mutually incomparable prototypes, each locally unimprovable for everyone at once, as archive-based multi-objective methods do \citep{laumanns2002}.

\subsection{How fast?}
\label{sec:rates}

Almost-sure convergence says nothing about time, which has two components. Discovery is governed by $\delta$: if tests were exact, the expected time to a local optimum would be at most the expected number of improvements on the way divided by $\delta$, a bound in the spirit of drift analysis \citep{heyao2001}. Recognition is governed by the noise. Telling apart two prototypes whose utilities differ by $\Delta$ takes a number of tests of the order of $\sigma^2/\Delta^2$ on each \citep{evendar2006, mannor2004}. With $\sigma=0.1$ and a difference of one hundredth of a utility unit, one comparison needs hundreds of tests, so the efficient response is to concentrate tests on the leading alternatives \citep{chen2000}.

\subsection{Stopping and reframing}
\label{sec:reframe}

Because the limit is local, more iteration cannot repair a poor frame, and a team must decide when to stop and when to reframe. The value of another round within the frame is at most the \emph{expected improvement}
\begin{equation}
\label{eq:ei}
\mathrm{EI}_k=\max_p\E\big[(U(p)-X_k)^+\mid\filt_k\big],
\end{equation}
the criterion of Bayesian optimisation \citep{jones1998, frazier2008}, taken over tested prototypes and unbuilt ideas. If a round costs $c_{\mathrm{iter}}$ in the same units, iterating is worthwhile at most while $\mathrm{EI}_k>c_{\mathrm{iter}}$. A reframe costs $c_{\mathrm{ref}}$ and offers an uncertain gain $G_{\mathrm{ref}}$ over what the team holds, and a one-step comparison suggests reframing when
\begin{equation}
\label{eq:reframe}
\mathrm{EI}_k-c_{\mathrm{iter}}<G_{\mathrm{ref}}-c_{\mathrm{ref}} .
\end{equation}
The right side contains the one quantity that cannot be derived from inside the frame. It can be estimated as an expected improvement one level up. If what the team comes to hold in a frame is regarded as a normal draw with mean $m$ and standard deviation $s$, learned from earlier projects, then
\begin{equation}
\label{eq:gref}
G_{\mathrm{ref}}=(m-X_k)\,\Phi(z)+s\,\phi(z),\qquad z=(m-X_k)/s,
\end{equation}
with $\Phi$ and $\phi$ the standard normal distribution and density functions. Frames are then the arms of a bandit, and \eqref{eq:reframe} is the greedy policy for it. It is a heuristic, and a lopsided one, since it sets the gain of a single round against that of a whole epoch; optimal search over alternatives has the reservation-price structure of \citet{weitzman1979}. Section~\ref{sec:reframesim} compares it with simpler rules.

\section{Simulation study}
\label{sec:sim}

\subsection{Setup}
\label{sec:setup}

A prototype is a subset of 14 features, so there are 16{,}384 prototypes, few enough to compute the utility of every one and to know the optimum $U^*$ and all local optima exactly. We use three classes of landscape. The \emph{main landscape} is an instance of \eqref{eq:utility} with weights $(w_{\mathrm d},w_{\mathrm f},w_{\mathrm v})=(0.45,0.25,0.30)$: desirability rewards the coverage of user needs with diminishing returns, feasibility falls with build and integration costs, and viability rewards business value and penalises scope. Its optimum is $0.763$, its mean utility $0.670$ and the standard deviation of utility across prototypes $0.046$; it has four local optima for one-feature variants and two for two-feature variants. The \emph{DVF family}, named after the three criteria, consists of 30 landscapes of the same form with independently drawn parameters (one to six local optima for one-feature variants). The \emph{NK family} consists of 20 landscapes from Kauffman's NK model \citep{kauffman1993}, in which each feature's contribution depends on 1, 2, 4 or 6 others; these are far more rugged (1 to 185 local optima, median 38).

A run starts from five random ideas. In each round the team picks one of its three \emph{leading prototypes}, those with the highest beliefs, and proposes five variants that differ from it in one or two features. Among all the ideas it has not yet built, it builds the one that looks best on a rough estimate and tests it once. It then re-tests its three leading prototypes and one further prototype, taking all prototypes in rotation for this last test so that each is tested again and again, as Assumption~\ref{as:test} requires, though at lengthening intervals; after 80 rounds about a third of the prototypes have been tested only once. A round uses five tests. Tests follow \eqref{eq:test} with $\sigma=0.10$ and no bias. Beliefs are conjugate normal posterior means under an uninformative prior (mean $0.5$, variance $0.08$), confined to $[0,1]$. The rough estimate used for pre-screening is the true utility plus noise of standard deviation $0.10$, drawn once per idea; it is as informative as one test and costs nothing, an advantage that Sections~\ref{sec:ablation} and~\ref{sec:baselines} remove.

Two choices shape what follows. The noise of a test is about twice the spread of utilities across the landscape, so a single test cannot tell an average prototype from a good one; we regard this as a fair picture of an early test with a handful of users, and Section~\ref{sec:ablation} halves and doubles it. A round is a substantial unit of work, so we report results from 5 rounds upward and treat 80 rounds, about 400 tests, as a long project. Utilities are reported as a percentage of $U^*$, a difference of one \emph{point} is one percent of $U^*$, and intervals are 95\% confidence intervals, computed across landscapes for the two families.

\subsection{The first 80 rounds}
\label{sec:dynamics}

\begin{figure}[t]
\centering
\includegraphics[width=0.6\textwidth]{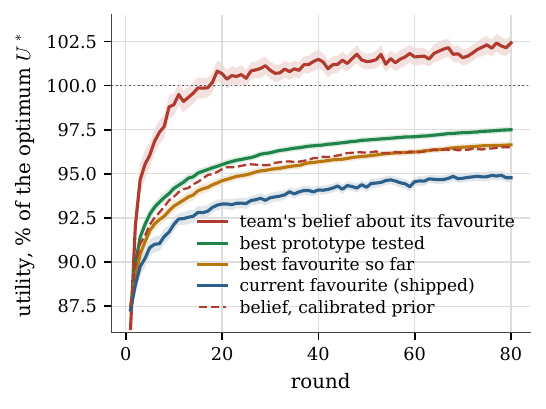}
\caption{The baseline loop on the main landscape (250 runs; bands are 95\% confidence intervals). The team's belief about its favourite exceeds the optimum from round 18 on, while the favourite is worth less than the best prototype the team has tested. A prior calibrated to the landscape (dashed) removes most of the overestimate.}
\label{fig:dynamics}
\end{figure}

On the main landscape (Figure~\ref{fig:dynamics}) the shipped utility is 90.8\% of the optimum after 5 rounds, 92.1\% after 10, 93.3\% after 20 and 94.8\% after 80. At each of these horizons the team has already tested something better than what it would ship: the best prototype found is worth 92.7\%, 94.2\%, 95.5\% and 97.5\%. After 80 rounds the team loses 2.5 points to prototypes it has not found and 2.7 points to its own choice among those it has. On the DVF family the two losses are 1.4 and 2.9 points, and on the rugged NK family, where good prototypes are harder to find, 3.6 and 2.7. The cost of choosing is thus comparable to the cost of not finding on every class, and larger on the smoother landscapes.

The team's belief about its favourite passes 100\% of the optimum at round 18 and ends at 102.4\%: the team believes it holds something better than the best prototype that exists. The overestimate after 80 rounds is $0.058\pm0.004$ in utility units, 7.6\% of the optimum; it is 7.3\% on the DVF family and 3.8\% on the NK family. This is the optimiser's curse of Section~\ref{sec:theory}. A prior matched to the main landscape cuts the overestimate to $0.014\pm0.002$ and leaves the shipped utility where it was (94.6\%): calibration corrects what the team believes about its result without improving the result.

\subsection{How far is the limit?}
\label{sec:horizon}

\begin{figure}[t]
\centering
\includegraphics[width=\textwidth]{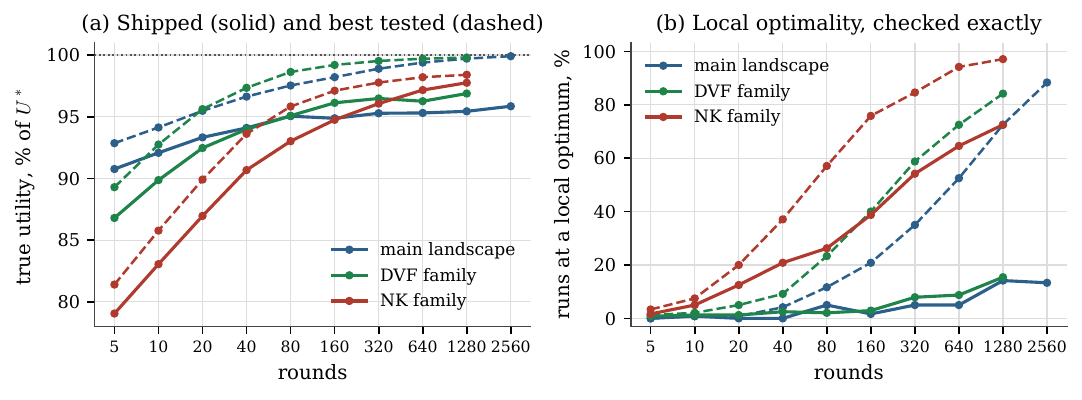}
\caption{Short and long runs (main landscape: 120 runs; DVF and NK families: 10 landscapes each, 24 runs per landscape). (a) Utility shipped (solid) and best utility among tested prototypes (dashed). (b) Share of runs in which the favourite (solid) or the best tested prototype (dashed) is a local optimum, determined exactly by enumeration.}
\label{fig:horizon}
\end{figure}

We ran the loop for up to 2{,}560 rounds and checked at each horizon whether the favourite is a local optimum, by looking it up in the enumerated set (Figure~\ref{fig:horizon}). On the main landscape it is one in 6 of 120 runs after 80 rounds and in 16 of 120 after 2{,}560 rounds, about 12{,}800 tests in a space of 16{,}384 prototypes. With one-feature variants the share after 2{,}560 rounds is 19\%. On the DVF family the share is 2\% after 80 rounds and 15\% after 1{,}280. On the NK family it is 26\% and 72\%, because four of the ten NK landscapes in this experiment have about twenty or more local optima even for two-feature variants, so that one is never far away.

A team that stops after a few dozen rounds has in most cases not converged. The obstacle is the ratio of test noise to the differences between prototypes, far more than the number of peaks. After 2{,}560 rounds on the main landscape the best prototype tested is a local optimum in 88\% of runs and is worth 99.9\% of the optimum, while the favourite is worth 95.8\%, so the gap between found and shipped has grown from 2.5 to 4.1 points. Late in a run the team holds thousands of prototypes within a few percent of each other, most of them tested once or twice, and the highest of several thousand noisy estimates rarely belongs to the best. Where prototypes differ more, as on the NK family, the gap closes, from 2.8 points after 80 rounds to 0.6 after 1{,}280.

One remedy is to discount thin evidence. Shipping the prototype with the highest belief minus one posterior standard deviation gains 0.4 points after 80 rounds on the main landscape and 2.4 after 2{,}560; after 1{,}280 rounds it gains 1.2 points on the DVF family and 0.2 on the NK family.

\subsection{New prototypes or re-tests?}
\label{sec:allocation}

\begin{figure}[t]
\centering
\includegraphics[width=\textwidth]{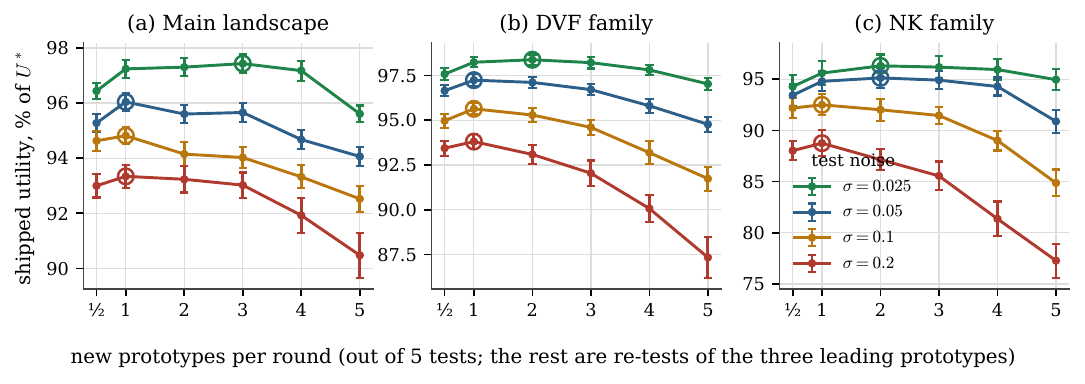}
\caption{Shipped utility after 60 rounds of exactly five tests each, as a function of how many of the five tests go to new prototypes, for four levels of test noise (five ideas per round; error bars are 95\% confidence intervals; circles mark the best allocation).}
\label{fig:allocation}
\end{figure}

We gave every team exactly five tests in each of 60 rounds and varied how many of the five went to new prototypes, from one every other round to all five. The remaining tests were re-tests of the three leading prototypes, so that the arms differ only in the use of tests (Figure~\ref{fig:allocation}; 120 runs on the main landscape and 16 on each landscape of the families, with 5 and with 15 ideas per round).

The curves rise and then fall. Giving all five tests to new prototypes, so that nothing is tested twice, is the worst allocation in 23 of the 24 combinations of landscape class, noise level and number of ideas. Against the best allocation it costs 1.6 to 2.9 points on the main landscape, 1.4 to 6.5 on the DVF family and 1.3 to 11.5 on the NK family, more as the noise grows. Building a new prototype only every second round is never the best allocation either. The best is one new prototype per round in 15 of the 24 combinations, two in five and three in four, and it moves towards re-testing as tests get noisier: with precise tests ($\sigma=0.025$) two or three new prototypes per round are best, and at the baseline noise and above one is best on every class. Tripling the number of ideas per round changes none of this.

A team should therefore build something new in every round and, unless its tests are very precise, spend most of its tests on prototypes it has tested before. As in the experiment of \citet{dow2010}, the gain comes from having alternatives to compare; the simulations add that the comparison is unreliable until the alternatives have been tested more than once.

\subsection{Relaxing the assumptions}
\label{sec:ablation}

Table~\ref{tab:ablation} changes one element of the baseline at a time.

\begin{table}[t]
\caption{One change at a time: utility shipped after 80 rounds, as a percentage of $U^*$, with 95\% confidence half-widths (main landscape: 200 runs; families: 16 runs on each landscape). The baseline uses variants within two features, five ideas per round, full retention, one new prototype, three re-tests of leading prototypes and one rotating re-test per round, and $\sigma=0.10$.}
\label{tab:ablation}
\centering\footnotesize\setlength{\tabcolsep}{4pt}
\begin{tabular}{@{}llccc@{}}
\toprule
Assumption & Changed element & Main & DVF family & NK family\\
\midrule
Baseline & --- & $94.9\pm0.3$ & $95.7\pm0.5$ & $93.7\pm0.9$\\
\addlinespace[2pt]
Local ideation & variants within 1 feature & $94.5\pm0.3$ & $95.2\pm0.4$ & $92.5\pm1.1$\\
 & variants within 3 features & $94.7\pm0.3$ & $95.9\pm0.4$ & $93.3\pm0.8$\\
 & unbounded variants & $94.2\pm0.3$ & $94.9\pm0.5$ & $92.4\pm0.8$\\
\addlinespace[2pt]
Generativity & 1 idea per round & $94.5\pm0.3$ & $94.4\pm0.5$ & $89.6\pm1.2$\\
 & 2 ideas per round & $94.8\pm0.3$ & $95.2\pm0.4$ & $92.0\pm0.9$\\
 & 15 ideas per round & $94.8\pm0.3$ & $95.8\pm0.4$ & $94.0\pm0.9$\\
 & ideas built without pre-screening & $94.0\pm0.3$ & $94.3\pm0.6$ & $89.0\pm1.4$\\
\addlinespace[2pt]
Retention & only the favourite kept & $93.7\pm0.3$ & $93.8\pm0.6$ & $90.1\pm1.0$\\
 & best 3 kept & $94.9\pm0.2$ & $95.7\pm0.4$ & $94.0\pm0.9$\\
 & best 10 kept & $94.9\pm0.3$ & $96.0\pm0.4$ & $93.4\pm0.9$\\
\addlinespace[2pt]
Testing & all five tests on new prototypes & $92.6\pm0.4$ & $92.0\pm0.7$ & $85.4\pm1.2$\\
 & no rotating re-test & $94.7\pm0.3$ & $95.7\pm0.4$ & $93.0\pm0.9$\\
 & noise $\sigma=0.05$ & $95.9\pm0.2$ & $97.5\pm0.4$ & $95.4\pm1.0$\\
 & noise $\sigma=0.20$ & $93.5\pm0.3$ & $93.6\pm0.5$ & $88.6\pm1.3$\\
\bottomrule
\end{tabular}
\end{table}

The reach of ideation has little effect within limits. One-feature variants cost 0.4 points on the main landscape and the DVF family and 1.2 on the NK family, three-feature variants change nothing measurable, and removing the bound altogether costs 0.7 to 1.3 points: unbounded ideation reaches the global optimum in the limit, but within 80 rounds it spends tests on distant prototypes that are mostly poor. The supply of ideas is important on rugged landscapes and saturates quickly elsewhere. One idea per round costs 0.4 points on the main landscape and 4.0 on the NK family, and fifteen are no better than five. Which idea is built counts for more: building one of the current round's ideas at random, in place of the idea that looks best among all those not yet built, costs 0.9, 1.4 and 4.7 points on the three classes. A team that remembers only its current favourite loses 1.2 to 3.6 points, and remembering three prototypes recovers all of it, so the favourite needs rivals to be compared with but not a complete archive.

Testing has the largest effects. Spending all five tests of a round on new prototypes costs 2.3, 3.7 and 8.2 points, although such a team finds prototypes as good as the baseline team does; it cannot tell which they are. Dropping only the rotating re-test, which the theorem needs for beliefs to become correct, costs at most 0.6 points within 80 rounds. Halving the noise of a test is worth 1.0 to 1.8 points and doubling it costs 1.4 to 5.1.

\begin{figure}[t]
\centering
\includegraphics[width=\textwidth]{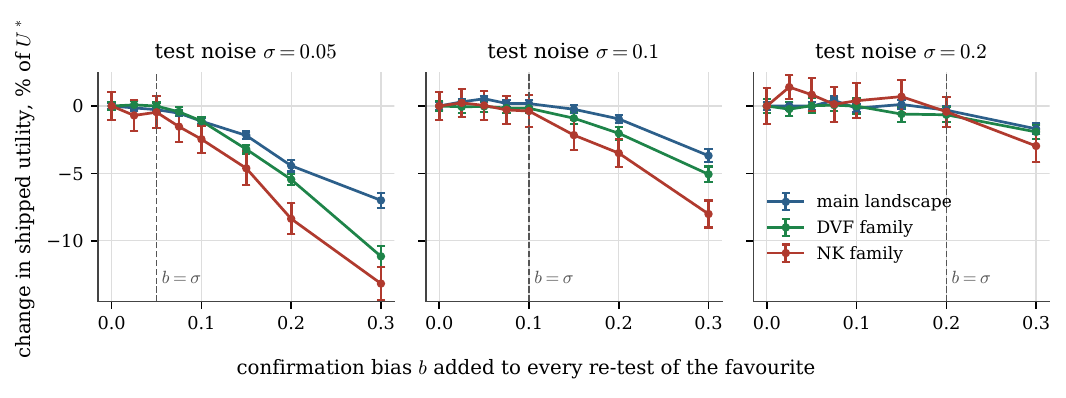}
\caption{Change in shipped utility after 80 rounds when each round's re-test of the favourite is inflated by $b$, at three levels of test noise (main landscape: 240 runs per point; families: 16 runs per landscape; error bars are 95\% confidence intervals). The dashed line marks $b=\sigma$.}
\label{fig:bias}
\end{figure}

Figure~\ref{fig:bias} adds a confirmation bias: in each round the re-test of the current favourite is inflated by a constant $b$. At the baseline noise of $0.10$ no loss is detectable on any class up to $b=0.10$; at $b=0.20$ the loss is 1.0 to 3.5 points and at $b=0.30$ it is 3.7 to 8.0. The point at which losses begin moves with the noise. With more precise tests ($\sigma=0.05$) losses are visible from $b=0.075$ and reach 1.1 to 2.5 points at $b=0.10$, and with noisier tests ($\sigma=0.20$) nothing is lost up to $b=0.20$. A bias changes the outcome only when it is large enough to reverse comparisons that the evidence would otherwise have decided, which here means a bias about as large as the standard deviation of a test. A team with precise tests is the more exposed to its own bias, because less noise is left in which a challenger can be lucky. The guarantee of Proposition~\ref{prop:bias} weakens in proportion to the bias, whereas the typical loss has a threshold.

\subsection{Comparison with other policies}
\label{sec:baselines}

We compared the loop with five alternatives given 300 tests, the same variants, the same test noise and the same beliefs (Table~\ref{tab:baselines}): random search among variants, hill-climbing over one-feature variants with repeated tests, an $\varepsilon$-greedy policy that either tests a new variant of the favourite or re-tests the favourite, an upper-confidence-bound (UCB) policy \citep{auer2002} and Thompson sampling \citep{russo2018}, the last two adding a new variant of the favourite at regular intervals. Because the loop screens its ideas and these policies do not, we also ran the loop without pre-screening and the UCB policy with it. The parameters of each alternative were tuned on ten DVF and eight NK landscapes, and all policies were scored on landscapes and seeds not used for tuning. The loop ran with its baseline settings.

\begin{table}[t]
\caption{Utility shipped after 300 tests, as a percentage of $U^*$, on landscapes not used for tuning (main landscape: 250 runs; families: 20 landscapes each, 30 runs per landscape; $\pm$ is the 95\% confidence half-width). ``Diff.'' is the mean paired difference from the loop across landscapes.}
\label{tab:baselines}
\centering\footnotesize\setlength{\tabcolsep}{4pt}
\begin{tabular}{@{}lccccc@{}}
\toprule
& Main & \multicolumn{2}{c}{DVF family} & \multicolumn{2}{c}{NK family}\\
\cmidrule(lr){3-4}\cmidrule(lr){5-6}
Policy & shipped & shipped & diff. & shipped & diff.\\
\midrule
Random search & $91.0\pm0.4$ & $87.9\pm0.8$ & $-7.4\pm0.8$ & $80.2\pm0.9$ & $-12.0\pm0.9$\\
Hill-climbing & $91.0\pm0.5$ & $88.7\pm0.8$ & $-6.6\pm0.8$ & $83.8\pm1.4$ & $-8.4\pm1.6$\\
$\varepsilon$-greedy & $94.0\pm0.3$ & $94.3\pm0.6$ & $-1.1\pm0.3$ & $89.9\pm1.1$ & $-2.3\pm0.6$\\
UCB & $94.5\pm0.3$ & $94.7\pm0.5$ & $-0.6\pm0.3$ & $90.2\pm1.1$ & $-2.1\pm0.5$\\
Thompson sampling & $93.9\pm0.3$ & $94.4\pm0.7$ & $-0.9\pm0.4$ & $89.6\pm1.1$ & $-2.7\pm0.6$\\
\addlinespace[2pt]
UCB with pre-screening & $95.0\pm0.2$ & $96.1\pm0.4$ & $+0.8\pm0.2$ & $93.6\pm1.0$ & $+1.4\pm0.5$\\
Loop without pre-screening & $93.8\pm0.3$ & $93.9\pm0.6$ & $-1.5\pm0.5$ & $87.5\pm1.2$ & $-4.7\pm0.7$\\
Loop & $94.7\pm0.3$ & $95.3\pm0.4$ & --- & $92.2\pm0.9$ & ---\\
\bottomrule
\end{tabular}
\end{table}

Random search and hill-climbing are 4 to 12 points behind the loop, because a policy that does not concentrate its tests on promising prototypes wastes most of them. Among the policies that do, the differences are much smaller, and the loop is not the best. Without pre-screening it is level with $\varepsilon$-greedy and Thompson sampling and 0.7 points behind UCB on the main landscape, 0.4 to 0.9 points behind the three bandit policies on the DVF family and 2.1 to 2.7 points behind them on the NK family. With pre-screening on both sides, UCB leads the loop by 0.3 points on the main landscape, which is within noise, and by 0.8 and 1.4 points on the families, where it wins on nine landscapes in ten. Pre-screening is worth 0.9, 1.5 and 4.7 points to the loop and 0.5, 1.4 and 3.5 points to UCB, which on every class exceeds the difference between the two policies. The loop's fixed routine is thus a good way to spend a testing budget, and the pre-screening of ideas contributes more than the choice of routine.

\subsection{Reframing}
\label{sec:reframesim}

A world contains six frames, each a different DVF landscape. A team has 150 rounds. It works in one frame at a time with the loop of Section~\ref{sec:setup}, without the rotating re-test, and a reframe costs a number of rounds and moves it to a frame it has not visited. When the rounds are spent, the team ships whichever of the favourites of the frames it has visited it believes to be best, and the payoff is that prototype's true utility. We compare a team that never reframes, teams that leave each frame after a fixed number of rounds, and teams that follow criterion~\eqref{eq:reframe} after a minimum stay of ten rounds. The criterion uses the team's beliefs only. A round costs $0.002$ in utility units and a reframe that many times its length, and $G_{\mathrm{ref}}$ follows \eqref{eq:gref} under three priors for what a frame yields: a default (mean $0.75$, standard deviation $0.05$), one calibrated to the beliefs teams hold after 75 rounds in these frames ($0.74$ and $0.04$), and an optimistic one ($0.79$ and $0.04$), each updated when the team leaves a frame. Every policy sees the same 240 worlds at each of four costs.

\begin{table}[t]
\caption{True utility of the prototype shipped after 150 rounds in a world of six frames, by reframing policy and by the cost of a reframe in rounds (240 paired runs; 95\% half-widths are at most $0.004$). The last column is the mean number of reframes per run at a cost of 10 rounds; the criterion rows differ in the prior used for $G_{\mathrm{ref}}$. Bold marks the best policy at each cost.}
\label{tab:reframe}
\centering\small
\begin{tabular}{@{}lccccc@{}}
\toprule
& \multicolumn{4}{c}{Cost of a reframe (rounds)} & \\
\cmidrule(lr){2-5}
Policy & 4 & 10 & 20 & 35 & Reframes\\
\midrule
Never & 0.694 & 0.694 & 0.694 & 0.694 & 0\\
\addlinespace[2pt]
Every 15 rounds & 0.700 & 0.697 & 0.693 & 0.694 & 5.0\\
Every 25 rounds & \textbf{0.703} & 0.698 & 0.698 & 0.698 & 4.0\\
Every 40 rounds & 0.702 & \textbf{0.704} & \textbf{0.703} & 0.698 & 2.0\\
Once, after 75 rounds & 0.701 & 0.702 & 0.699 & 0.699 & 1.0\\
\addlinespace[2pt]
Criterion, default & 0.703 & 0.701 & 0.694 & 0.699 & 1.2\\
Criterion, calibrated & 0.702 & 0.702 & 0.697 & \textbf{0.699} & 1.2\\
Criterion, optimistic & 0.703 & 0.701 & 0.693 & 0.695 & 1.9\\
\bottomrule
\end{tabular}
\end{table}

A single reframe at the midpoint is better than none at every cost, clearly so when a reframe is cheap (Table~\ref{tab:reframe}). The paired gain is $0.007\pm0.004$ and $0.008\pm0.004$ in utility at costs of 4 and 10 rounds and $0.004\pm0.004$ and $0.005\pm0.004$ at costs of 20 and 35, about one percent of what the team ships. Reframing more often adds little: leaving a frame every 40 rounds is better than the single reframe by $0.001$ to $0.005$ at costs up to 20 rounds, mostly within the confidence intervals, and leaving every 15 rounds is no better than never reframing once a reframe costs 10 rounds or more.

The criterion does about as well as the single scheduled reframe and no better. With the calibrated prior it differs from it by less than the confidence interval at all four costs, and with the default prior at three of them. In most runs it fires at the earliest permitted moment, so it works as an early exit from frames that look poor against the prior and seldom as a signal that iteration has run dry. The expected improvement of a further round does not decline along an epoch: new, thinly tested rivals keep it at an average of $0.011$ both in rounds 10 to 20 and in rounds 40 to 75, above the cost of a round in 86\% of all rounds. Most of the value of reframing thus lies in doing it once, and criterion~\eqref{eq:reframe} adds nothing to a fixed plan as an operating rule.

\section{Discussion}
\label{sec:discussion}

\subsection{When design thinking works}

In the limit, the iterate--test loop works when the five assumptions hold, and each of them names a practice: keep the brief fixed while iterating, stay within reach of what has been built, keep producing variants, test in ways that can contradict the team and test favourites again, and keep what has been tried. When these hold, improvement follows from the structure of the process, which is consistent with Liedtka's proposal that the method works by disciplining judgement \citep{liedtka2015}. Table~\ref{tab:conditions} sets the assumptions beside what a departure from each cost in the simulations.

\begin{table}[!htbp]
\caption{The five assumptions of Theorem~\ref{thm:converge}: what each means in practice, what the theory says when it fails, and what a departure in its direction cost after 80 rounds (range over the three landscape classes, in points of $U^*$; from Table~\ref{tab:ablation} and Figure~\ref{fig:bias}).}
\label{tab:conditions}
\centering\small
\begin{tabularx}{\textwidth}{@{}p{2.4cm}XXp{3.9cm}@{}}
\toprule
Assumption & In practice & When it fails & Simulated departure and its cost\\
\midrule
Finite frame & The brief and the vocabulary of features stay fixed during an epoch & No fixed space in which an optimum is defined & Not simulated\\
Local ideation & New ideas are variants of what has been built & With unbounded jumps the limit is global but progress is slower & Unbounded variants: $-0.7$ to $-1.3$\\
Persistent generativity & The team keeps proposing variants, even late & The process can stop short of a local optimum & One idea per round in place of five: $-0.4$ to $-4.0$\\
Unbiased, repeated testing & Tests can contradict the team; favourites are tested again & A lucky prototype can stay on top; under bias $B$ the result is only $2B$-locally optimal & No prototype tested twice: $-2.3$ to $-8.2$. Bias $b=0.2$: $-1.0$ to $-3.5$\\
Retention & Rivals to the favourite are kept & The best prototype found can be lost & Only the favourite kept: $-1.2$ to $-3.6$\\
\bottomrule
\end{tabularx}
\end{table}

At the horizon of an actual project the answer shifts. Design thinking is usually presented as a discipline of generating: many ideas, many cheap prototypes, an early move from talking to building \citep{brown2008, dschool2010}. In our simulations generation is only half of the difficulty. At every horizon teams have tested something better than what they would ship, their confidence in what they would ship is inflated, and a prototype tested once is a poor basis for a decision. The pattern that an emphasis on building invites, many different prototypes each shown to users once, is the worst allocation we found. Experiments on idea selection point the same way: groups that generate more ideas select no better \citep{rietzschel2006}, and the quality of a group's best idea depends among other things on its ability to discern quality \citep{girotra2010}.

Four recommendations follow. A team should screen its ideas before building, since a rough estimate of which idea is most promising was worth more in our comparisons than the difference between the loop's fixed routine and a tuned bandit policy. It should treat its estimate of its favourite concept as an overestimate; the optimiser's curse grows with the number of alternatives chosen among \citep{smithwinkler2006}. It should build something new in every round and give most of its remaining tests to its two or three leading concepts, preferring a well-tested concept to one that looked excellent once. And it should worry about confirmation bias most when its tests are precise. On reframing the simulations support a plain rule: plan for at least one return to the problem definition, and do not expect iteration to signal when it is due.

For tools the implication is that evaluation, more than generation, is the scarce resource. A tool that supported it would keep a belief and an evidence count for every concept tested and shrink estimates that rest on little evidence. The same reasoning applies to generative AI as a source of ideas. In a story-writing experiment, ideas from a language model raised the rated creativity of individual stories and made the stories more alike \citep{doshi2024}. In the model more ideas per round help only on rugged landscapes, and a language model would be more useful if it could rank ideas roughly before any is built.

\FloatBarrier
\subsection{Limitations}

The model is normative: it describes a team that keeps beliefs and revises them consistently, which real teams do informally at best. The frame is taken as given. The model is silent on how a point of view is formed and on what makes one frame better than another, which is where much of the skill in design lies and where problems are wicked in the sense of \citet{rittel1973} and \citet{buchanan1992}. Holding the frame fixed during an epoch also sits uneasily with the view that problem and solution evolve together; Corollary~\ref{cor:drift} accommodates a problem that shifts within bounds and nothing more. 

Within the frame, the simplification with the largest consequences concerns what a test returns. In the model a test returns a number. In practice it also returns reasons: a user test shows why a prototype fails, and the reasons direct the next idea. The model therefore understates what a single test of a new prototype is worth, and the allocation result should be read with that in mind. Independent beliefs work in the same direction, since a test of one prototype teaches something about its relatives; correlated beliefs would speed up recognition and call for the correlated knowledge gradient \citep{frazier2009}. A prototype is, moreover, a set of features drawn from a fixed list, which leaves out changes of architecture or of a continuous parameter.

The simulations use synthetic landscapes with fourteen features. The qualitative results hold across three classes of very different ruggedness, but we have not varied the size of the space, and the magnitudes are not estimates for real projects. The threshold for bias was located at three noise levels only, the reframing experiment uses one class of landscape, and pre-screening, which is free in our simulations, would cost something in practice.

\section{Testing the model on real teams}
\label{sec:empirical}

In the vocabulary of practice a prototype is a concept, and we use that word here. Any account with noisy evaluation predicts that teams sometimes carry forward a concept that is not their best. Three more specific consequences would distinguish the model. Re-testing should reduce the cost of choosing: the shortfall of the concept a team carries forward, relative to the best concept it tested, should be larger in teams that tested each concept once than in teams that returned to their leading concepts, and a team's own assessment of its chosen concept should exceed an independent assessment by more when it chose among more candidates. Allocation should have an interior optimum: among teams with similar testing effort, outcomes should be best at an intermediate share of tests spent on new concepts and should fall off sharply as that share approaches one. And bias should have a threshold: leading questions and tests confined to the favourite should be associated with worse outcomes only where the bias they introduce is large relative to the variability of test results. The prediction that teams which reframe do better cannot be tested on observational records, because teams reframe when they are in trouble; it requires an experiment in which a reframe is scheduled for some teams and not for others.

The quantities involved can be obtained from recorded design sessions by protocol analysis \citep{cross1996}. A session is segmented into moves \citep{goldschmidt2014}, each labelled as proposing a concept, building one, testing one, or returning to the problem definition, and for each test the coder records the concept, the direction and strength of the evidence, and markers of likely bias, with agreement between coders reported by a chance-corrected measure \citep{krippendorff2018}.

A study by \citet{sung2019} shows both that such coding supports a sequential analysis and where its limits lie. They coded nine think-aloud sessions, in which triads of fourth-grade pupils worked on a paper design task, into seven cognitive strategies (699 coded events, with 85\% raw agreement on four sessions coded twice) and analysed the 505 transitions by lag-sequential analysis \citep{bakeman2011}. Generating an idea (their code \emph{Designing}) was followed by anticipating its consequences (\emph{Predicting}) more often than chance would give, in 33 of the 147 transitions out of generating ($z=3.66$), and predicting was followed by a return to generating in 39 of 52 cases ($z=5.17$). This cycle takes place in thought and is close to what we call pre-screening. The pupils built and tested nothing and the sessions were pooled, so the study bears on the feasibility of the measurement and not on the loop of building and testing.

In a confirmatory study the unit of analysis is the framing epoch, because the model's claims hold within a frame. The outcome must be measured independently of the team's own beliefs, by judges who rate every concept the team tested without knowing which one it chose; the consensual assessment technique gives a procedure and a reliability check for such ratings \citep{amabile1982}, and since the ratings are themselves noisy, several judges are needed for each concept. Comparisons should be made within epochs where possible. For a paired comparison with a standardised difference of 0.3, about one hundred epochs give a power of 0.9 at a one-sided level of 5\%.

\section{Conclusion}

Within a fixed framing of a problem the loop of ideation, prototyping and testing converges to a locally optimal prototype when the frame is finite, ideation is local, improving ideas keep coming at a non-summable rate, tests are unbiased and repeated, and tested prototypes are retained. The guarantee weakens in proportion to a bounded bias or drift of the criteria, and with several stakeholders it becomes convergence to a locally Pareto-optimal set.

The simulations show that the limit describes the direction of the process more than the position of a team that stops. Except on very rugged landscapes few runs reach a local optimum, and the result depends as much on how well the team can choose among what it has built as on what it has found. A prototype tested once is a poor basis for a decision, the team's confidence in its favourite is too high, confirmation bias does harm once it exceeds the noise in the tests, and one reframe captures most of what reframing has to offer.

Design thinking works, in the sense made precise here, when a team keeps its options, generates near them, screens what it builds, and tests its leading prototypes more than once. Whether teams that do these things obtain better outcomes is a question for field data, and Section~\ref{sec:empirical} says which data would settle it.

\section*{Data availability}
The code that produces every figure, table and number in this paper, together with the raw outputs, is openly available in a public repository at \url{https://github.com/zxhouse/dt-formalisation/}. The experiments use fixed random seeds and require only \texttt{numpy} and \texttt{matplotlib}.

\section*{Conflict of interest}
The authors declare that they have no known competing financial interests or personal relationships that could have appeared to influence the work reported in this paper.

\section*{Acknowledgements}
The authors thank participants of seminars and workshops for helpful comments and discussions on earlier drafts of this work.

\section*{Funding}
This research did not receive any specific grant from funding agencies in the public, commercial, or not-for-profit sectors.

\section*{CRediT authorship contribution statement}
\textbf{Pawe{\l} Kura{\'s}}: Conceptualization, Methodology, Formal analysis, Software, Writing -- original draft.
\textbf{Adrian Michalski}: Formal analysis, Validation, Writing -- review \& editing.
\textbf{Alicja Gerka}: Conceptualization, Investigation, Writing -- review \& editing.
\textbf{Patryk Organi{\'s}ciak}: Formal analysis, Validation, Writing -- review \& editing.

\end{document}